\documentclass[letterpaper, 10 pt]{ieeeconf}  %

\usepackage{comment}

\usepackage{epsfig}
\usepackage{graphicx}
\usepackage{amsmath}
\usepackage{amssymb}
\usepackage{multirow}
\usepackage{multicol}
\usepackage{colortbl}
\usepackage[usenames,dvipsnames]{xcolor}
\usepackage{xcolor,colortbl}
\usepackage{xfrac}
\usepackage{dblfloatfix}
\usepackage{flushend}
\usepackage{verbatim}
\usepackage{booktabs}
\usepackage{adjustbox}
\usepackage{wrapfig}
\usepackage{microtype}
\usepackage{makecell}
\usepackage{bbm}
\usepackage{cite}
\usepackage{microtype}      %

\usepackage{floatrow}

\usepackage{xcolor}
\usepackage{textcomp}
\usepackage{color}
\usepackage{colortbl}
\usepackage{url}

\usepackage{graphicx}

\usepackage{subfig}
\usepackage{tikz}
\usepackage{pgfplots}
\usepackage{tikz-3dplot}
\tdplotsetmaincoords{60}{115}
\pgfplotsset{compat=newest}
\DeclareMathOperator{\E}{\mathbb{E}}
\usepackage[colorlinks = true]{hyperref}

\title{\huge The Distracting Control Suite -- A Challenging Benchmark for Reinforcement Learning from Pixels}

\author{
  Austin Stone$^{*}$ \hspace{1cm} Oscar Ramirez \hspace{1cm} Kurt Konolige \hspace{1cm} Rico Jonschkowski$^{*}$ \\
  Robotics at Google, \texttt{\small \{austinstone,oars,konolige,rjon\}@google.com}, $^{*}$equal contribution
}

\begin{document}
\maketitle
\thispagestyle{empty}
\pagestyle{empty}

\begin{abstract}
Robots have to face challenging perceptual settings, including changes in viewpoint, lighting, and background. Current simulated reinforcement learning (RL) benchmarks such as DM Control~\cite{deepmindcontrolsuite2018} provide visual input without such complexity, which limits the transfer of well-performing methods to the real world. In this paper, we extend DM Control with three kinds of visual distractions (variations in background, color, and camera pose) to produce a new challenging benchmark for vision-based control, and we analyze state of the art RL algorithms in these settings. 
Our experiments show that current RL methods for vision-based control perform poorly under distractions, and that their performance decreases with increasing distraction complexity, showing that new methods are needed to cope with the visual complexities of the real world.  We also find that combinations of multiple distraction types are more difficult than a mere combination of their individual effects. 
\end{abstract}

\newcommand\w{0.15\linewidth}

\section{Introduction}
\label{sec:introduction}

The DeepMind Control Suite (DM Control)~\cite{deepmindcontrolsuite2018} is one of the main benchmarks for continuous control in the reinforcement learning (RL) community. By providing a challenging set of tasks with a fixed implementation and a simple interface, it has enabled a number of advances in RL -- most recently a set of methods that solve the benchmark as well and efficiently from pixels as from states~\cite{CuRL,DRQ,RAD}. Simulation-based benchmarks like DM Control have many advantages: they are easy to distribute, they are hermetic and repeatable, and they are fast to train and iterate on.  However, the DeepMind Control Suite is a poor proxy for real robot learning from visual input, which remains inefficient despite the advances we have seen in DM Control. To enable research gains in simulated benchmarks to better translate to gains in real world vision-based control, we need a new simulated benchmark that more closely mirrors perceptual challenges of real environments, most importantly visual \emph{distractions} -- variations in the input that are irrelevant for the task.

A major challenge in perception is to extract only the task-relevant information from sensory input and remove distractions which might otherwise lead to spurious correlations in downstream tasks~\cite{jonschkowski2015learning,zhang2020learning}. DM Control does not contain such distractions, as the agent is shown from a constant camera view under constant lighting against a singular, static background. Since every change in the observation is tied to the change of a task-relevant state variable, DM Control does not allow measuring or making progress on the ability of filtering out irrelevant variations through perception.

\begin{figure*}[t]
\centering
\includegraphics{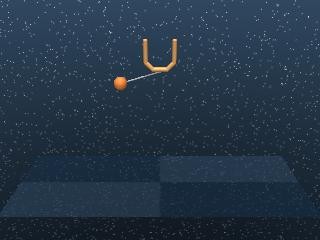}
\includegraphics{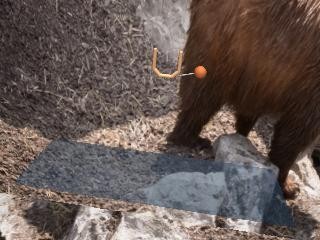}
\includegraphics{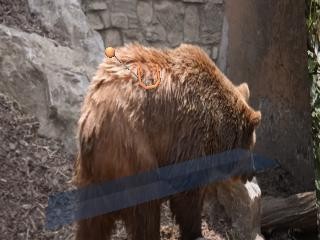}
\includegraphics{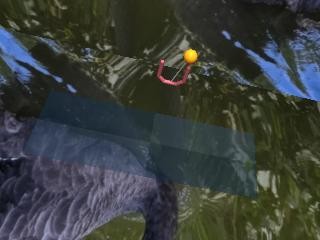}
\includegraphics{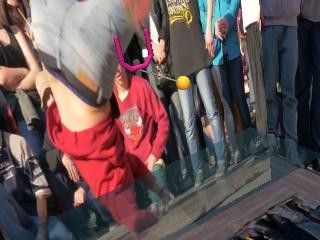}
\includegraphics{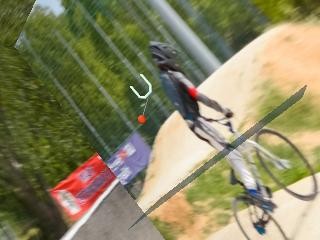}
\includegraphics{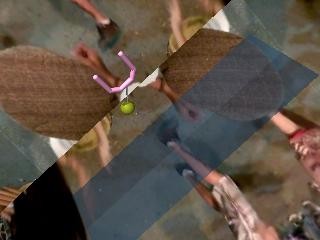}
\includegraphics{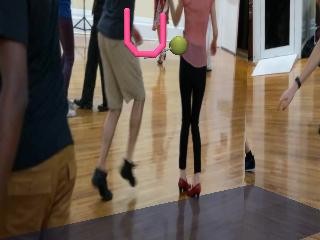}
\includegraphics{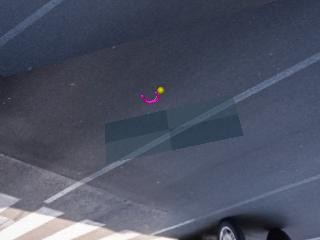}
\includegraphics{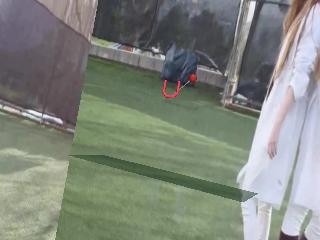}
\includegraphics{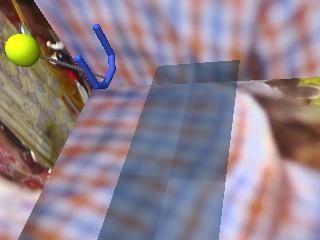}
\\
\includegraphics{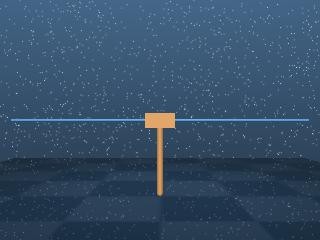}
\includegraphics{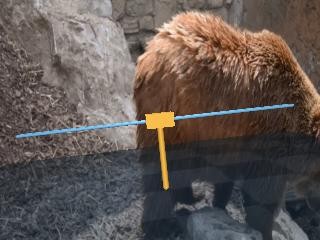}
\includegraphics{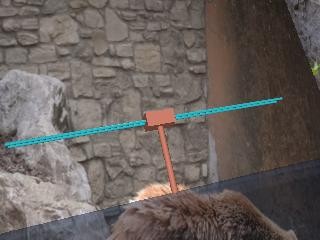}
\includegraphics{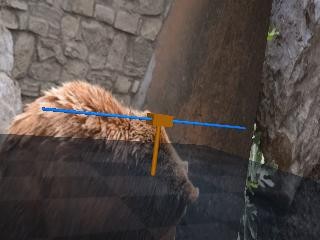}
\includegraphics{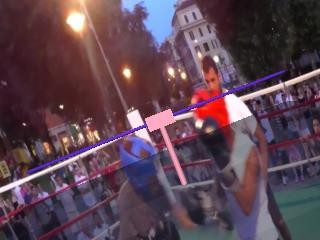}
\includegraphics{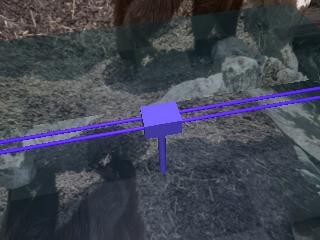}
\includegraphics{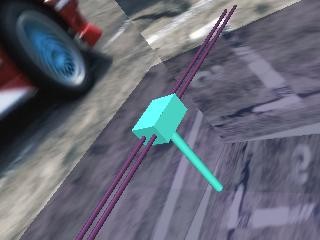}
\includegraphics{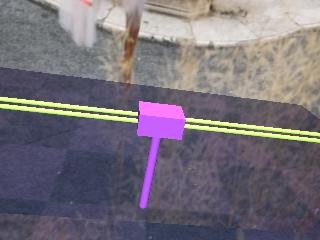}
\includegraphics{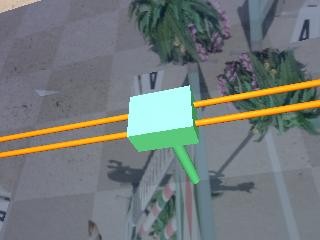}
\includegraphics{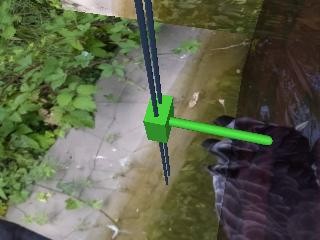}
\includegraphics{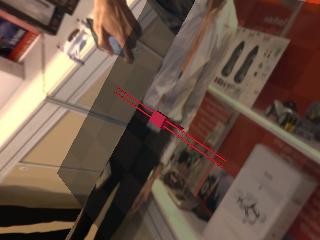}
\\
\includegraphics{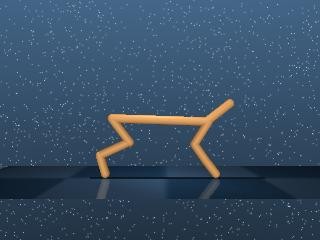}
\includegraphics{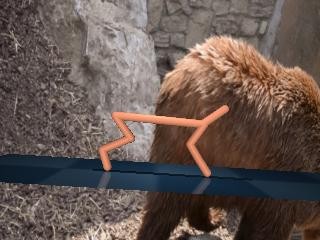}
\includegraphics{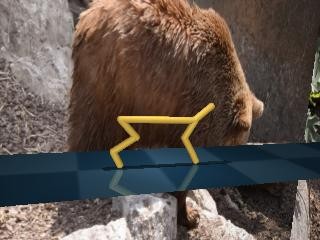}
\includegraphics{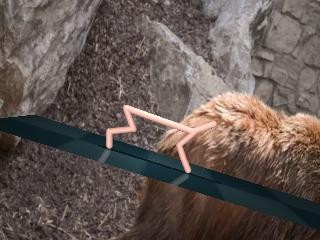}
\includegraphics{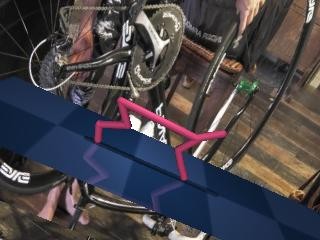}
\includegraphics{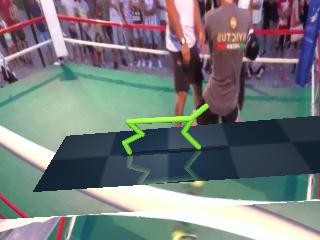}
\includegraphics{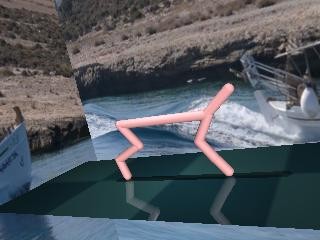}
\includegraphics{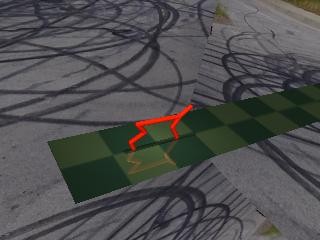}
\includegraphics{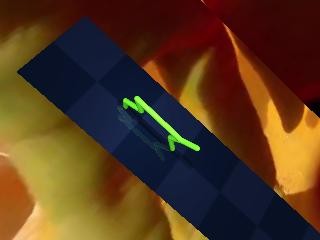}
\includegraphics{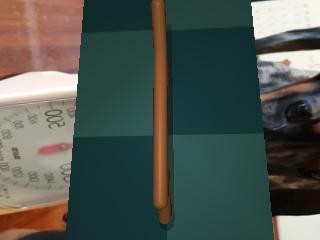}
\includegraphics{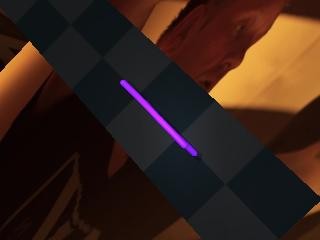}
\\
\includegraphics{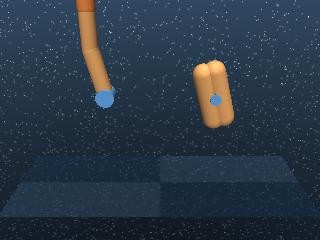}
\includegraphics{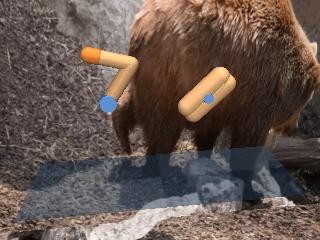}
\includegraphics{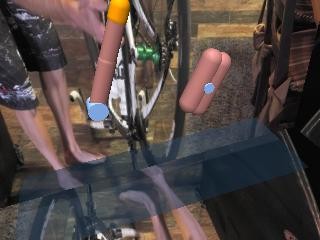}
\includegraphics{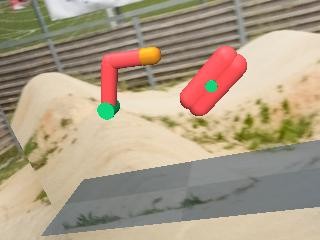}
\includegraphics{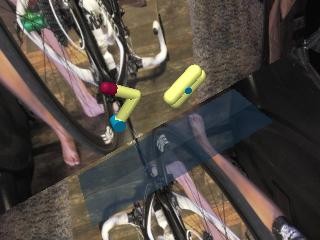}
\includegraphics{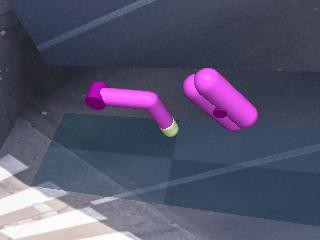}
\includegraphics{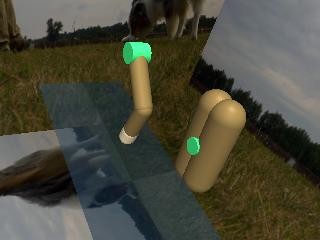}
\includegraphics{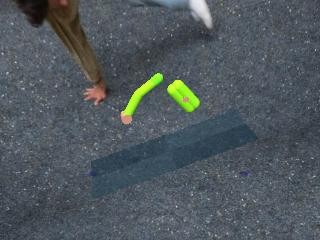}
\includegraphics{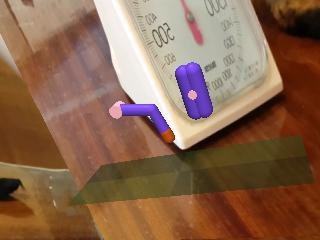}
\includegraphics{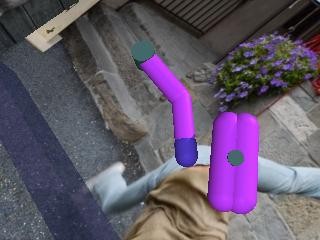}
\includegraphics{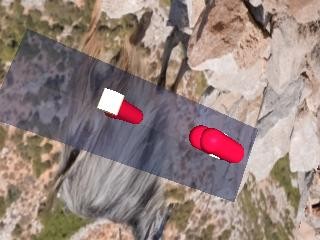}
\\
\includegraphics{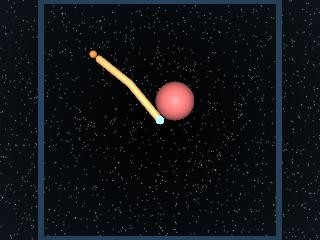}
\includegraphics{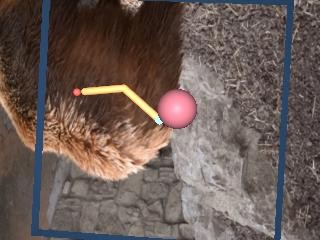}
\includegraphics{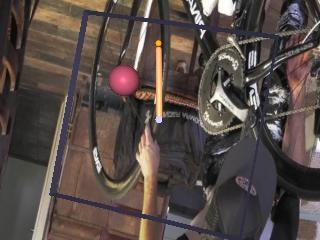}
\includegraphics{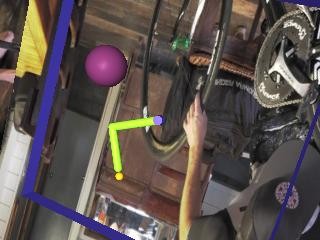}
\includegraphics{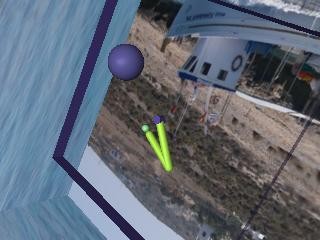}
\includegraphics{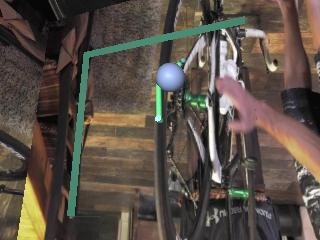}
\includegraphics{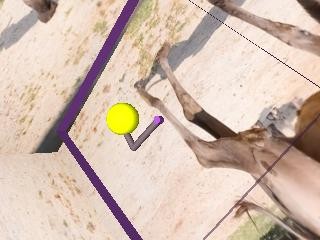}
\includegraphics{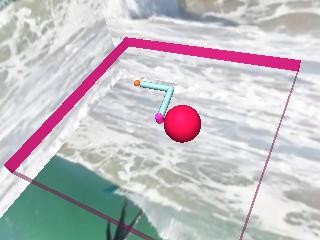}
\includegraphics{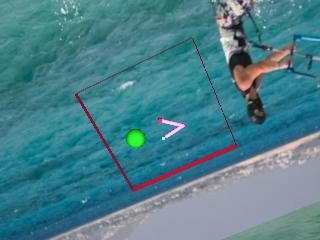}
\includegraphics{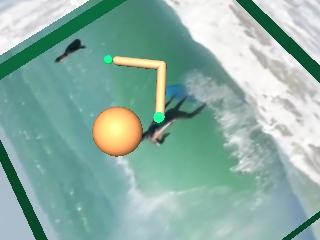}
\includegraphics{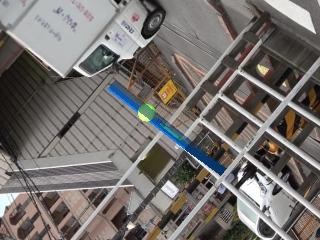}
\\
\includegraphics{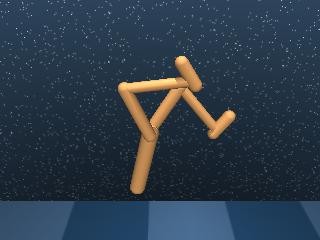}
\includegraphics{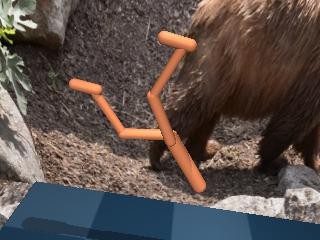}
\includegraphics{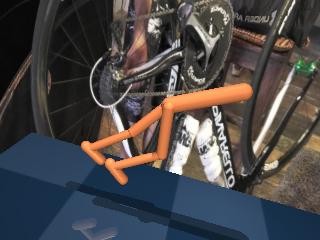}
\includegraphics{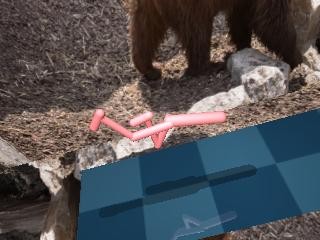}
\includegraphics{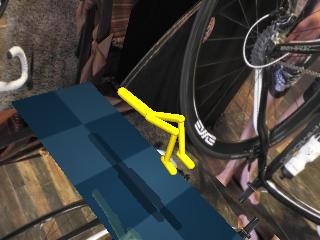}
\includegraphics{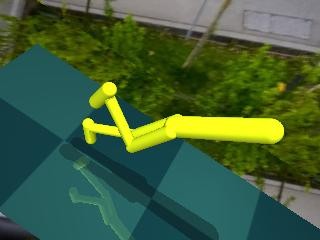}
\includegraphics{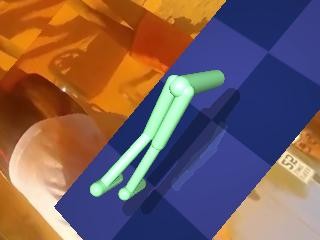}
\includegraphics{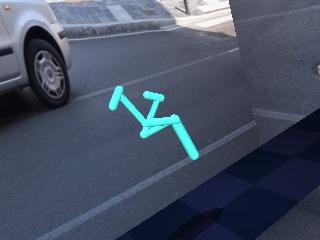}
\includegraphics{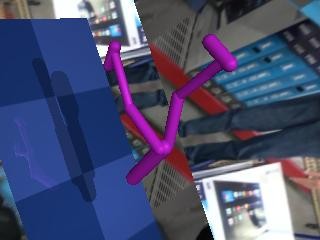}
\includegraphics{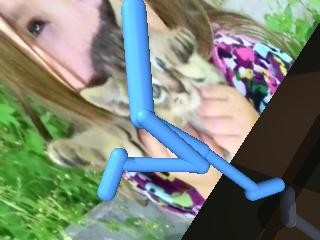}
\includegraphics{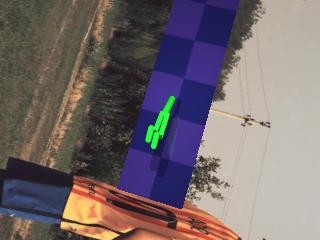}
\\\vspace{-0.10cm}
\caption{The Distracting Control Suite. The six tasks (one per row) are shown at increasing levels of difficulty (columns). From left to right, camera and color distractors are shown in 0.1 increments from 0 to 1. The number of backgrounds per column is increased from 0 to 1 and then doubles at each column after that up to a maximum of 60. The first column shows the \emph{no distractions} benchmark. The second column showcases the \emph{easy} benchmark on one of the 4 available background videos. The third column is our \emph{medium} benchmark. Current state-of-the-art methods stop learning effective policies at this point.\vspace{-0.3cm}}
\label{fig:example}
\end{figure*}

To address this problem, we present the \emph{Distracting Control Suite}, an extension of DM Control created with real-world robot learning in mind. Our extension adds three distinct types of distractions: random color changes of all objects in the scene, random video backgrounds, and random continuous changes of the camera pose (see Fig.~\ref{fig:example}). Each of these distractions can be applied in a \emph{static} setting where changes only occur at episode transitions or in a \emph{dynamic} setting where distractions change smoothly between frames. All distractions can be scaled in their difficulty from barely perceptible to severely distracting. All three distraction types can be arbitrarily combined with each other.

We implement these distractions on top of DM Control to retain the same simple interface. Our suite works by accessing and modifying scene properties (color, camera position, and background textures) at run time before visual observations are rendered. The underlying physics and control properties of the tasks are kept exactly the same to facilitate comparisons to work performed on the original DM Control.

\setkeys{Gin}{width=0.085\linewidth}

Using the Distracting Control Suite, we perform an empirical analysis of state of the art methods in reinforcement learning from pixels, comparing different combinations of SAC~\cite{SAC} and QT-Opt~\cite{QT} with RAD~\cite{RAD} and DrQ~\cite{DRQ}. We analyze a) the sensitivity to distractions during inference when no distractions are present during training, b) the effect of each individual distraction on RL performance at different difficulties, and c) a combination of all three distractions which proves to be challenging for existing methods.

We have three main contributions: 1) The design and implementation of the Distracting Control Suite, which is available at \url{https://github.com/google-research/google-research/tree/master/distracting_control} and which we hope will facilitate future advances in vision-based control. 2) The definition of a benchmark with results for the current state of the art that future work can compare against. 3) A set of empirical observations about RL from pixels when faced with distractions, such as i) methods are relatively robust to especially color distractions without training on them but struggle to improve substantially from seeing distractions during training, ii) distractions interact in a way that makes combinations of them especially difficult, iii) the relative performance of different methods changes significantly between the DM Control and our Distracting Control benchmarks. We think that these observations are especially relevant to real world, robot RL where task-irrelevant visual input is very common. We hope that our benchmark can be a useful proxy for learning visual control in the real world and therefore facilitate advances in robot learning.

\section{Related Work}
\label{sec:related_work}

Learning successful policies from pixels in the Atari environment~\cite{DQN} was a major breakthrough in reinforcement learning that produced a surge of interest and advances in pixel-based RL. The work in simulation first focused on Atari, but later also included DM Control from pixels~\cite{pmlr-v97-hafner19a}. Recently, CURL~\cite{CuRL}, DrQ~\cite{DRQ}, and RAD~\cite{RAD} have established that different versions of applying image cropping augmentation can greatly improve results up to a point where DM Control can be solved similarly from pixels as from states. Reinforcement learning has also been successfully applied to robotics training in the real world ~\cite{pmlr-v28-levine13, QT, SACX, InfiniteReplay}. 

An alternative approach to training on real robots is to use \emph{domain randomization} to train in a very diverse set of simulated environments that enables transfer to the real world~\cite{SadeghiL16,tobin2017domain,rubikscube}. Domain randomization is the extension of data augmentation, which has been used in computer vision since the inception of convolutional networks~\cite{lecun1998gradient}, from data sets to simulators. Randomizing many aspects of the simulation that do not match the real world forces the learned model to be robust to these variations.

\emph{Distractions}, which this paper focuses on, can look technically similar to domain randomization but distractions as we define them here are part of the problem that the agent has to solve rather than part of the solution. As a result, the agent does not have control over distractions, i.e. cannot affect these distractions, cannot arbitrarily sample more of them, and has to handle them during evaluation. The importance of visual distractions for studying perception and control was first demonstrated in simple environments~\cite{jonschkowski2015learning} and has recently been applied to more complex ones \cite{zhang2018natural,antonova2020benchmarking,zhang2020learning,Hansen2020}, including different modifications to the DeepMind Control Suite~\cite{deepmindcontrolsuite2018}.

The goal of our work is to provide a unifying benchmark with visual distractions to enable comparability between approaches for pixel-based RL that currently rely on different sets of distractions. Compared to distractions that were added to DM Control in previous or concurrent work, our benchmark combines camera, color, and background distractions, and presents an in-depth study of state of the art methods in this new setting. We hope that our empirical observations and our Distracting Control Suite with clearly defined benchmarks will facilitate future research in this direction.

\section{The Distracting Control Suite}
\label{sec:suite}

This work extends the DeepMind Control Suite~\cite{deepmindcontrolsuite2018} to make its perception aspect more challenging by adding visual distractions. The resulting \emph{Distracting Control Suite} applies random changes to camera pose, object colors, and background. The magnitude of each distraction type can be controlled by a ``difficulty magnitude" scalar between 0 and 1. Distractions can be set to either change during episodes or change only between episodes, which we will refer to as \emph{dynamic} and \emph{static} settings, respectively.

For the viewing camera, the difficulty magnitude scales both the span of camera poses and the camera velocity. For the color change augmentations, the difficulty magnitude scales the maximum allowable color change and the speed of color changes, and for the background distractors it scales the number of unique videos used or (for one of our experiments) the weight for blending between the background videos and the original skybox background.

\subsection{Camera Pose}
\label{sec:suite_camera}

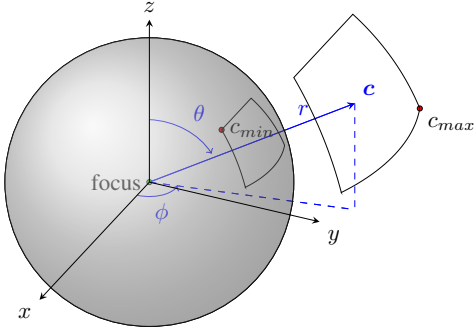
\begin{figure}
\hfill
\adjustbox{width=0.75\linewidth}{
\begin{tikzpicture}[tdplot_main_coords, scale = 2.2]
  \def\rvec{1.}
  \def\thetavec{70}
  \def\phivec{110}
  \def\outer{2.}
  
  \def\thetamid{60}
  \def\phimid{100}
  \def\rmid{1.7}
  
  \def\thetamin{45}
  \def\phimin{70}
  \def\outermax{2.}
  \tdplotsetcoord{O} {0}{0}{0}
  \tdplotsetcoord{P}{\outer}{\thetavec}{\phivec}
  \tdplotsetcoord{A}{\rvec}{\thetamin}{\phimin}
  \tdplotsetcoord{B}{\rmid}{\thetamid}{\phimid}
  \tdplotsetcoord{C}{\outer}{\thetavec}{\phivec}
  \draw[-stealth,color=blue] (O) -- (B) node[near end,above] {$r$};
  \draw[fill = red] (P) circle (0.5pt) node[below right] {$c_{max}$};
  \draw[fill = red] (A) circle (0.5pt) node[right] {$c_{min}$};
  \draw[fill = green] (O) circle (0.5pt) node[left] {$\text{focus}$};
  \draw[dashed,blue]   (O)  -- (Bxy);
  
  \draw[-stealth,blue] (O)  -- (B) node[above right] {$\boldsymbol{c}$};
  \draw[dashed,blue]   (O)  -- (Bxy);
  \draw[dashed,blue]   (B)  -- (Bxy);

  \tdplotdrawarc[->, color=blue, near start]{(O)}{0.2}{0}{\phimid}
    {anchor=north}{$\phi$}
  \tdplotsetthetaplanecoords{\phivec}
  \tdplotdrawarc[->,tdplot_rotated_coords, color=blue]{(0,0,0)}{0.5}{0}{\thetamid}
    {anchor=south west}{$\theta$}

\pgfmathsetmacro{\Radius}{1}

 \draw[tdplot_screen_coords] (0,0) circle (\Radius);
 \begin{scope}
  \draw plot[variable=\x,domain=-20:20] (xyz spherical cs:radius=\Radius,latitude=20,longitude=\x)
  -- plot[variable=\x,domain=20:45] (xyz spherical cs:radius=\Radius,latitude=\x,longitude=20)
  -- plot[variable=\x,domain=45:20] (xyz spherical cs:radius=\Radius,latitude=\x,longitude=-20);
 \end{scope}
 \pgfmathsetmacro{\RadiusBig}{2.}
 \draw[tdplot_screen_coords] (0,0) circle (\Radius);
 \begin{scope}
  \draw plot[variable=\x,domain=-20:20] (xyz spherical cs:radius=\RadiusBig,latitude=20,longitude=\x)
  -- plot[variable=\x,domain=20:45] (xyz spherical cs:radius=\RadiusBig,latitude=\x,longitude=20)
  -- plot[variable=\x,domain=45:20] (xyz spherical cs:radius=\RadiusBig,latitude=\x,longitude=-20);
 \end{scope}
 
\shade[ball color = lightgray, opacity = 0.5] (0,0,0) circle (1cm);

\draw[-stealth] (0,0,0) -- (1.80,0,0) node[below left] {$x$};
\draw[-stealth] (0,0,0) -- (0,1.30,0) node[below right] {$y$};
\draw[-stealth] (0,0,0) -- (0,0,1.30) node[above] {$z$};
\end{tikzpicture}}
\vspace{-0.15cm}
\caption{Specification of camera pose range.\vspace{-0.3cm}}
\label{fig:cam}
\end{figure}

We parameterize the camera pose by $c = (\phi, \theta, r, \theta_{roll})$, corresponding to the spherical angles $\phi$ and $\theta$ and radius $r$, which define the camera position, and an additional angle $\theta_{roll}$ that specifies the roll. The camera's pitch and yaw are not randomly varied. Depending on whether the task uses a tracking camera, e.g. for \emph{cheetah} and \emph{walker}, or a ``fixed" camera, e.g. for \emph{cartpole}, pitch and yaw are calculated to focus on the agent's current or starting position, respectively. The difficulty scale defines a viewing range of the camera as a subset of the upper frontal hemisphere for azimuth and elevation that scales the maximum distance (see Fig.~\ref{fig:cam}).  Based on the difficulty scale $\beta_{\text{cam}}\in[0,1]$, we set $\phi_{max} = \theta_{max} = \theta_{roll\,max} = \frac{\pi \beta_{\text{cam}}}{2}$, $r_{min} = r_{\text{original}} (1 - 0.5 \beta_{\text{cam}})$, and $r_{max} = r_{\text{original}} (1 + 1.5 \beta_{\text{cam}})$. Therefore, $0 \leq \phi,\theta, \theta_{roll} \leq \frac{\pi}{2}$, and $0.5 r_{\text{original}} \leq r \leq 2.5 r_{\text{original}}$. In the \emph{static} setting, we uniformly sample the camera pose from this range at the start of each episode and keep it constant during the episode. In the \emph{dynamic} setting, we sample the camera's starting pose in the same way, but additionally maintain a camera velocity $v_{t}$ that is updated via a random walk at each time step.
$$
\begin{aligned}[c]
v_0&\thicksim \mathcal{U}(v_{min}, v_{max}),\\
v_n&=v_0 + \sum_{j=1}^n \mathcal{N}(0,\sigma \Sigma),
\end{aligned}
\hspace{1cm}
\begin{aligned}[c]
c_0&\thicksim \mathcal{U}(-c_{min}, c_{max}),\\
c_n&=c_0 + \sum_{j=1}^n v_n.
\end{aligned}    
$$
Velocity is stored as both an $(\dot x, \dot y, \dot z)$ spatial vector and a $\dot\theta$ roll velocity. The random walk's standard deviation and maximum velocity are scaled relative to the viewing range, $v_{max}=\frac{2\beta_{\text{cam}}}{5}$, $\sigma=\frac{\beta_{\text{cam}}}{10}$, $v_{roll\,max}=\frac{\pi\beta_{\text{cam}}}{50}$, $\sigma_{roll}=\frac{\pi\beta_{\text{cam}}}{300}$. The random walk is clipped to within the maximum velocity and camera pose parameters.

\subsection{Object Colors}
\label{sec:object_colors}

For this distraction type, we change the colors of all bodies in the simulation, where the difficulty scalar $\beta_\text{rgb}\in[0,1]$ defines the maximum distance per color channel. At the start of each episode, all colors are sampled uniformly per channel $x_0\thicksim \mathcal{U}(x - \beta_\text{rgb}, x + \beta_\text{rgb})$, where $x_0$ is a sampled color value and $x$ is the original color in DM Control. In the static setting, the colors remain constant throughout the episode. In the dynamic setting, they change randomly $x_n=x_{n-1} +  \mathcal{N}(0,0.03\cdot\beta_\text{rgb})$, but are clipped to never exceed the maximum distance $\beta_\text{rgb}$ from the original color.

\subsection{Background}
\label{sec:suite_background}

Here, we project random backgrounds from videos of the DAVIS 2017 dataset \cite{DAVIS} onto the skybox of the scene. To make these backgrounds visible for all tasks and views, we make the floor plane transparent except for walking tasks where it is small and task relevant (we set the ground plane opacity $\alpha=1.0$ for \emph{cheetah} and \emph{walker}, $\alpha=0$ for \emph{reacher}, $\alpha=0.3$ for all other tasks). Depending on the experiment, we use a different number of background videos $b\in[0,60]$ -- the task is more difficult when more scenes are used. We take the $b$ first videos in the DAVIS 2017 training set and randomly sample a video and a frame from it at the start of every episode. In the static setting, that frame stays constant. In the dynamic setting, the video plays forwards or backwards until the last or first frame is reached at which point the playing direction is reversed. This way, the background motion is always smooth and without ``cuts''. In one experiment, we smoothly blend between the distraction background and the original skybox background with weights $\beta_\text{bg}$ and $1-\beta_\text{bg}$ respectively (see Fig.~\ref{fig:blending}).

\section{Methods for RL from Pixels}
\label{sec:methods}

Our experiments compare SAC~\cite{SAC} and QT-Opt~\cite{QT} with and without random cropping following the RAD~\cite{RAD} approach with a single random cropping per sample or averaging over two crops as detailed in DrQ~\cite{DRQ}. This can also be viewed as using DrQ with $K=M\in\{0,1,2\}$. While QT-Opt in fact already includes random cropping in its description, for consistency, we will refer to that approach as QT-Opt+RAD and have QT-Opt denote the method without cropping.

To implement QT-Opt+DrQ, we modify the Bellman error minimization. Originally QT-Opt proposes
$$
\mathcal{E}(\theta) = \E_{(\mathbf{s}, \mathbf{a}, \mathbf{s'})\sim p(\mathbf{s}, \mathbf{a}, \mathbf{s'})}[D(Q_\theta(\mathbf{s},\mathbf{a}), Q_t(\mathbf{s}, \mathbf{a}, \mathbf{s'}))],
$$
where the cross-entropy function is used as the divergence metric $D$ and $Q_t$ is the \emph{target value} defined by $Q_t(\mathbf{s}, \mathbf{a}, \mathbf{s'})=r(\mathbf{s},\mathbf{a}) + \gamma V(\mathbf{s'})$. To compute $V$, QT-Opt estimates a $Q$-function and uses CEM~\cite{rubinstein1999cross} to select the best action according to the current $\hat{Q}$ estimate. Adding DrQ augmentations requires two changes to the algorithm. First, we need to average the target value over $K=2$ random image crops,
$$
\mathbf{y} = r(\mathbf{s},\mathbf{a}) + \frac{1}{K}\sum_{k=1}^{K}\gamma V_{\theta'}\left(f(\mathbf{s'}, \mathbf{\upsilon}_k)\right),
$$
where $f$ is the image transformation function and $\upsilon_k \sim \mathcal{U}$ a random sample of image augmentation parameters. Second, we need to average the $Q$ estimates in the loss,
$$
J_Q(\theta) = (\mathbf{y} - \frac{1}{M}\sum_{m=1}^{M}Q_{\theta}(f(\mathbf{s}), \mathbf{\upsilon}_m), \mathbf{a}))^2.
$$
All methods use faithful replications of their published hyperparameters without special tuning to the Distracting Control Suite. Note that while SAC+DrQ was originally tuned for DM Control from pixels, QT-Opt was only tuned for DM Control with state input.

\section{Experiments}
\label{sec:experiments}

In this section, we analyze state of the art reinforcement learning methods in our Distracting Control Suite, which yields a number of interesting results: 1) Methods trained without any distractions are fairly robust to color distractions an somewhat robust to camera distractions during inference. 2) Training with distractions does not substantially improve this robustness, except for background distractions where performance improves but only up to a point. For random color changes in particular, the improvement from training with these distractions is minor compared to not training with them. 3) Training with random video backgrounds performs better than training with random static backgrounds. Generalization to new backgrounds is limited and does not improve when training on additional background scenes. 4) The degrading effects of distractions on task performance are more than multiplicative. As a result, current methods perform rather poorly in our benchmarks that combine three kinds of distractions, even in the easiest settings. 5) The ranking of methods changes from the standard DM Control benchmark -- where SAC-based and QT-Opt-based methods perform comparably -- to our distracting benchmarks, where QT-Opt with RAD or DrQ augmentations performs best. Generally, we found that RAD variants worked equally well or better than DrQ across our experiments.

\paragraph*{Network Architecture}
All methods use the same model architecture from DRQ~\cite{DRQ}. A shared image encoder applies four convolutional layers using $3\times3$ kernels and 32 filters with an stride of 2 for the first layer and 1 for others. ReLU activations are applied after each convolution. A final 50 dimensional output dense layer normalized by LayerNorm~\cite{ba2016layer} is applied with a tanh activation. Both critic and actor networks (in the case of SAC) are parametrized with a 3-layer MLP using ReLU activations up until the last layer. The output dimension of these layers is 1024. In the critic this reduces to a single Q-Value prediction, and in the case of the actor it predicts a mean and covariance for each action. The image encoder weights are shared when using SAC across the critic and the actor, and gradients are only computed through the critic optimizer.

\begin{table*}[t]
\floatsetup{captionskip=0cm}
\begin{floatrow}
\ttabbox[0.45\linewidth]{\centering
    \adjustbox{width=\linewidth}{
    \begin{tabular}{lc}
    \toprule
    Task & AR \\
    \midrule
    Ball In Cup Catch & 4 \\
    Cartpole Swingup & 8 \\
    Cheetah Run & 4 \\
    Finger Spin & 2 \\
    Reacher Easy & 4 \\
    Walker Walk & 2\\
    \bottomrule
    \end{tabular}
    }}{\caption{Tasks and action repeats (ARs)}
    \label{table:action_repeats}}
\ttabbox[1.54\linewidth]{\centering
    \adjustbox{width=\linewidth}{
    \begin{tabular}{llllllll}
    \toprule
    Method & Mean & BiC-Catch & C-Swingup & C-Run & F-Spin & R-Easy & W-Walk\\
    \midrule
    SAC & 265$\pm$13 & 146$\pm$26 & 384$\pm$42 & 165$\pm$41 & 481$\pm$9 & 188$\pm$6 & 228$\pm$17 \\
    SAC+RAD & \bf 836$\pm$29 & 962$\pm$2 & 843$\pm$10 & \bf 515$\pm$13 & \bf 976$\pm$5 & 962$\pm$8 & \bf 762$\pm$184 \\
    SAC+DrQ & \bf 808$\pm$24 & 958$\pm$2 & \bf 859$\pm$6 & \bf 546$\pm$26 & 808$\pm$75 & 968$\pm$2 & \bf 711$\pm$171 \\
    QT-Opt & 372$\pm$18 & 418$\pm$81 & 425$\pm$26 & 218$\pm$8 & 600$\pm$38 & 306$\pm$21 & 264$\pm$30 \\
    QT-Opt+RAD & \bf 820$\pm$3 & \bf 968$\pm$1 & \bf 843$\pm$14 & \bf 538$\pm$11 & 953$\pm$1 & \bf 969$\pm$5 & \bf 648$\pm$25 \\
    QT-Opt+DrQ & \bf 801$\pm$5 & 962$\pm$2 & \bf 851$\pm$5 & \bf 534$\pm$12 & 952$\pm$1 & \bf 974$\pm$1 & \bf 532$\pm$29 \\
    \bottomrule
    \end{tabular}}}{\caption{DM Control results (without distractions) at 500K steps. Mean $\pm$ standard error. Highest mean scores and results that are not significantly different ($p<0.1$) are boldfaced.}
    \label{table:default}}
\end{floatrow}
\end{table*}

\paragraph*{Tasks and Experiment Parameters}

Training is performed with batch size 512, and alternates one learning step with each sample collection step. Tasks and action repeats are adopted from the Planet benchmark (see Table~\ref{table:action_repeats}). All experiments report results after 500K environment steps, evaluated for 100 episodes. Unless otherwise noted all experiments are performed with five random seeds per task used to compute means and standard errors of their evaluations. In tables, results are boldfaced if they have the highest mean or if they do not have a statistically significant difference ($p<0.1$) from the result with the highest mean.

\subsection{Robustness to Distractions During Inference} 
\label{eval_robust}

In this experiment, we analyze how well methods trained on the standard DM Control benchmark generalize to unseen distractions during inference. After training each method without any distractions, we then test them separately for each type of distraction with different amounts of distracting variation $\beta_\text{rgb}$, $\beta_\text{cam}$, $\beta_\text{bg}$ from 0 to 1. The number of background scenes $b=60$.

\begin{figure*}[t]
\centering
\includegraphics{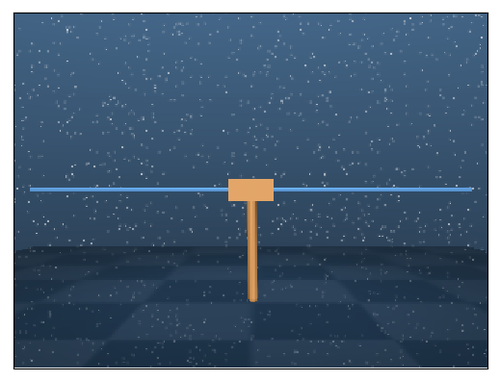}
\includegraphics{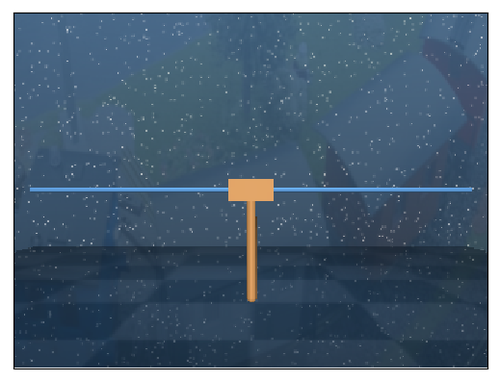}
\includegraphics{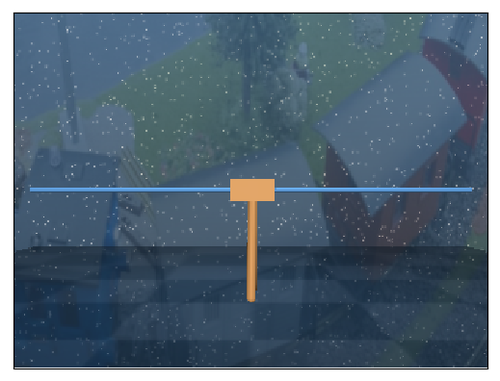}
\includegraphics{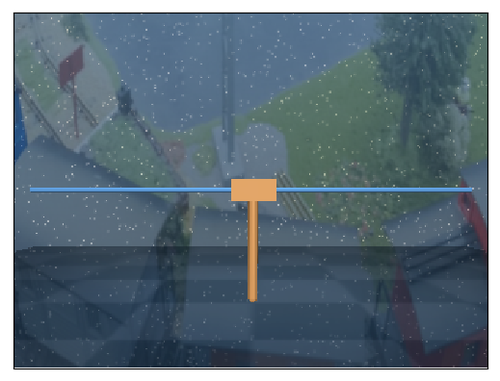}
\includegraphics{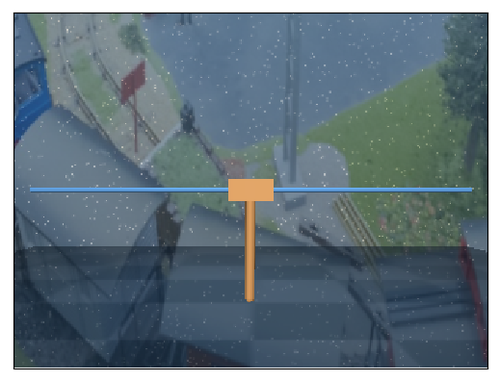}
\includegraphics{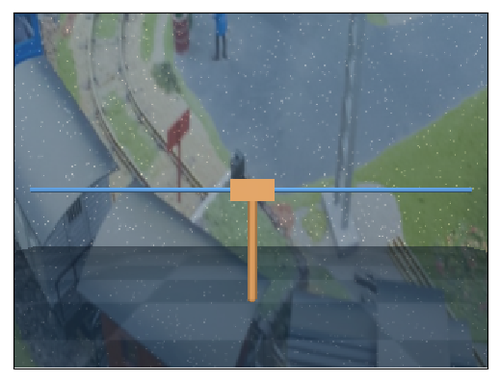}
\includegraphics{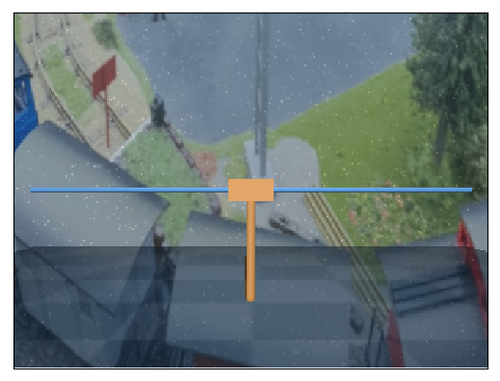}
\includegraphics{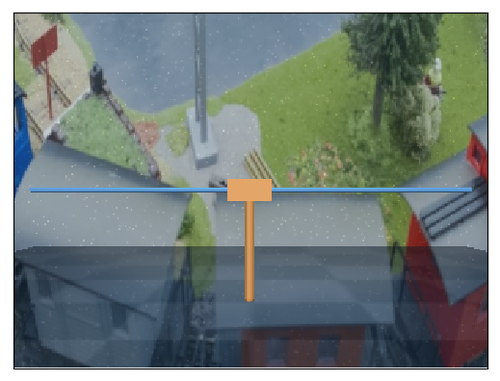}
\includegraphics{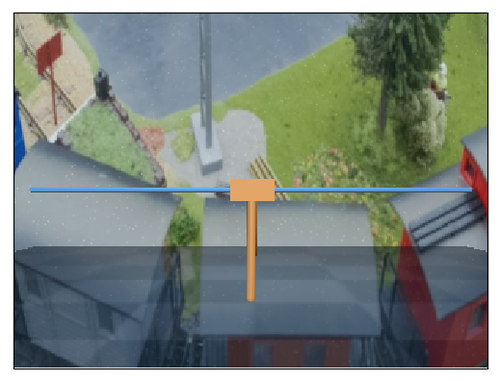}
\includegraphics{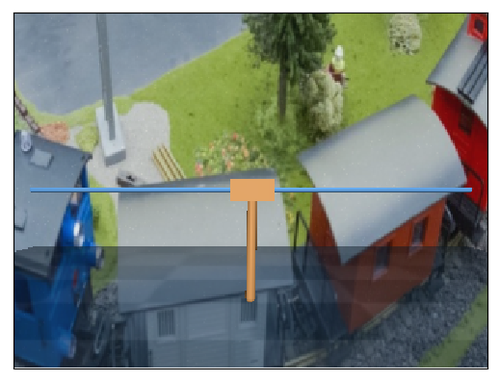}
\includegraphics{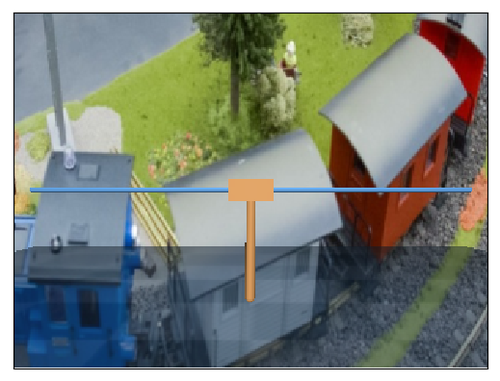}
\\\vspace{-0.1cm}
\caption{Blending between the original skybox and the distracting background with $\beta_\text{bg}\in[0,1]$.\vspace{-0.2cm}}
\label{fig:blending}
\end{figure*}

\begin{figure*}[t]
\centering
\subfloat[Camera]{\includegraphics[]{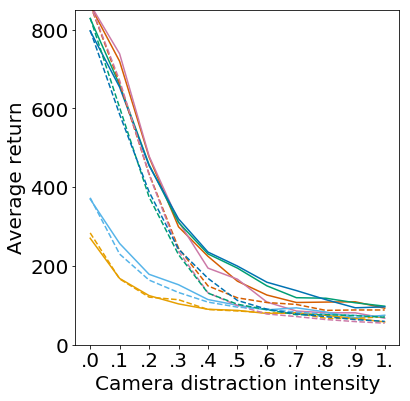}}
\hfill
\subfloat[Color]{\includegraphics[]{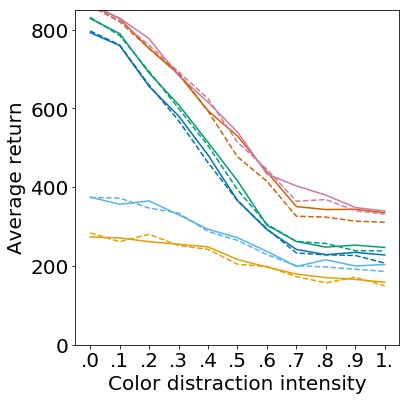}}
\hfill
\subfloat[Background]{\includegraphics[]{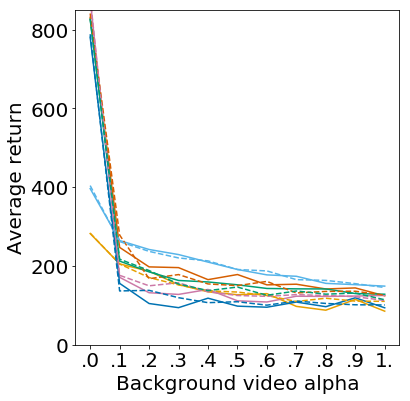}}
\hfill
\includegraphics[]{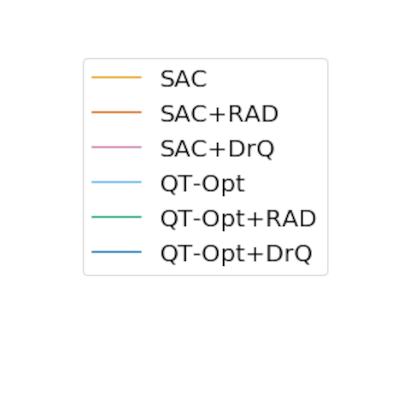}
\vspace{-0.1cm}
\caption{Evaluating with each distraction type after training without distractions. Distraction intensities $\in [0,1]$ (see Sect. \ref{sec:suite}). Lines show means over all 6 tasks. Colors denote methods, solid/dashed lines are results in the static/dynamic setting.}
\label{fig:eval_with_distractions}
\end{figure*}

Table~\ref{table:default} shows the results in DM Control without distractions and verifies that methods are learning to solve the tasks. We see that using one or two cropping augmentations (RAD or DrQ) is necessary for reaching high performance and that SAC-based methods and QT-Opt based methods perform comparably. Figure~\ref{fig:eval_with_distractions} evaluates these trained models with camera, color, and background distractions of different intensities. As expected, all methods lose performance with increasing distraction intensity, but the robustness to these distractions varies with the distraction type and method. All methods cope best with color distractions (b), less well with camera pose distractions (a), and are highly sensitive to unseen backgrounds even when blended with the skybox background (c, visualized in Fig.~\ref{fig:blending}). The points where the top methods lose half of their score are at camera scale $\beta_\text{cam}=0.2$ (corresponding to camera views in column 3 of Fig.~\ref{fig:example}), at color scale $\beta_\text{rgb}=0.6$ (corresponding to color changes in column 7 of Fig.~\ref{fig:example}), and at a background weight $\beta_\text{bg}<0.1$, which corresponds to column 2 in Fig.~\ref{fig:blending}. It seems to be irrelevant if the distractions are dynamic or static over an episode (dashed vs. solid lines). Interestingly, SAC-based methods appear more robust to color distractions than QT-Opt-based methods (b).

\setkeys{Gin}{width=0.085\linewidth}
\setkeys{Gin}{width=\linewidth, trim=0 0 0 0}

\setkeys{Gin}{width=0.245\linewidth, trim=0 0 0 0, clip}
\setkeys{Gin}{width=\linewidth, trim=0 0 0 0}

\setkeys{Gin}{width=0.245\linewidth, trim=0 0 0 0, clip}
\begin{figure*}[t]
\centering
\subfloat[Camera]{\includegraphics[]{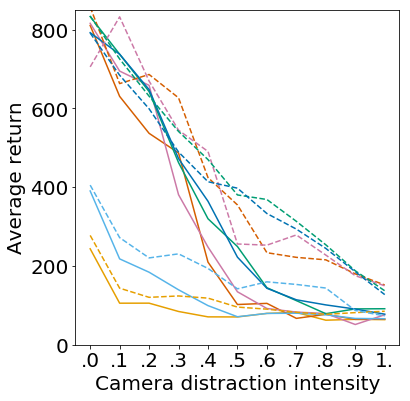}}
\hfill
\subfloat[Color]{\includegraphics[]{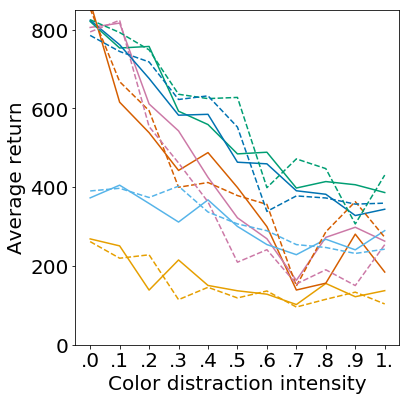}}
\hfill
\subfloat[Background]{\includegraphics[]{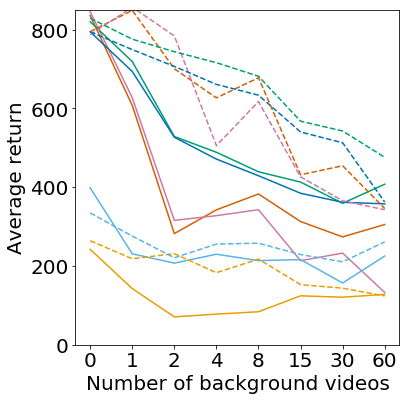}}
\hfill
\subfloat[Background (unseen)]{\includegraphics[]{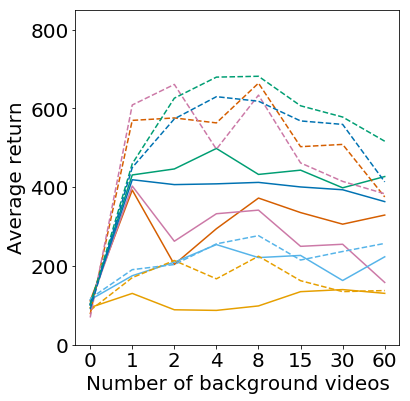}}
\vspace{-0.1cm}
\caption{Effect of distraction magnitude when distractions are present during training and evaluation. Same legend as above. \vspace{-0.2cm}}
\label{fig:train_with_distractions}
\end{figure*}
\setkeys{Gin}{width=\linewidth, trim=0 0 0 0}

\subsection{Training with Distractions}

\begin{table*}[t]
\CenterFloatBoxes
\floatsetup{captionskip=0cm}
\begin{floatrow}
\ttabbox[1.2\linewidth]{\centering
    \adjustbox{width=\linewidth}{
    \begin{tabular}{lllllllll}
    \toprule
    & Method & Mean & BiC-Catch & C-Swingup & C-Run & F-Spin & R-Easy & W-Walk\\
    \midrule
    \parbox[t]{1mm}{\multirow{6}{*}{\rotatebox[origin=c]{90}{Static setting}}} 
    & SAC & 94$\pm$4 & 104$\pm$17 & 211$\pm$7 & 64$\pm$8 & 52$\pm$15 & 82$\pm$10 & 49$\pm$14 \\
    & SAC+RAD & 182$\pm$24 & 129$\pm$20 & 360$\pm$25 & 72$\pm$44 & 370$\pm$114 & 102$\pm$14 & 60$\pm$31 \\
    & SAC+DrQ & 166$\pm$24 & 138$\pm$20 & 334$\pm$29 & 4$\pm$2 & 378$\pm$125 & 113$\pm$22 & 28$\pm$1 \\
    & QT-Opt & 149$\pm$7 & 81$\pm$20 & 215$\pm$3 & 118$\pm$5 & 198$\pm$23 & 132$\pm$11 & \bf 152$\pm$6 \\
    & QT-Opt+RAD & \bf 317$\pm$8 & \bf 218$\pm$44 & \bf 446$\pm$23 & \bf 220$\pm$5 & \bf 711$\pm$27 & \bf 181$\pm$17 & 128$\pm$14 \\
    & QT-Opt+DrQ & 299$\pm$6 & \bf 217$\pm$35 & \bf 416$\pm$20 & 199$\pm$8 & \bf 695$\pm$33 & \bf 171$\pm$25 & 93$\pm$9 \\
    \midrule
    \parbox[t]{1mm}{\multirow{6}{*}{\rotatebox[origin=c]{90}{Dynamic setting}}} 
    & SAC & 98$\pm$7 & 103$\pm$18 & 176$\pm$3 & 79$\pm$10 & 19$\pm$12 & 99$\pm$10 & \bf 110$\pm$22 \\
    & SAC+RAD & 270$\pm$31 & 366$\pm$59 & 297$\pm$21 & \bf 198$\pm$39 & \bf 338$\pm$59 & 173$\pm$11 & \bf 249$\pm$138 \\
    & SAC+DrQ & 199$\pm$30 & 247$\pm$41 & 235$\pm$12 & 92$\pm$37 & 238$\pm$58 & 221$\pm$12 & \bf 164$\pm$136 \\
    & QT-Opt & 118$\pm$5 & 72$\pm$25 & 172$\pm$1 & 88$\pm$7 & 86$\pm$12 & 137$\pm$21 & \bf 155$\pm$6 \\
    & QT-Opt+RAD & \bf 343$\pm$24 & \bf 490$\pm$64 & \bf 467$\pm$12 & \bf 170$\pm$8 & \bf 393$\pm$91 & \bf 428$\pm$68 & \bf 109$\pm$12 \\
    & QT-Opt+DrQ & 265$\pm$5 & \bf 395$\pm$39 & 431$\pm$18 & 126$\pm$10 & 203$\pm$33 & \bf 343$\pm$53 & \bf 91$\pm$3 \\
    \bottomrule
    \end{tabular}}
    }{%
    \caption{Benchmark easy, $\beta_{\text{cam}}=\beta_{\text{rgb}}=0.1$, $b=4$ background videos}\label{table:easy}
    }
\killfloatstyle\ffigbox[0.78\linewidth]{
    \centering
    \includegraphics[width=\linewidth]{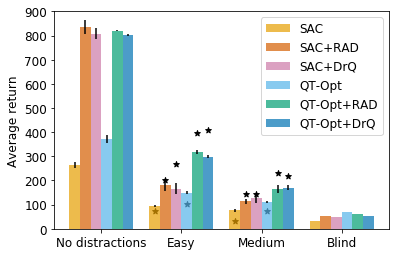}
}{%
    \caption{Benchmarks in the static setting averaged over all tasks. Bars show means and standard errors. Stars indicate expected scores if degradations were independent per distraction.}
    \label{fig:benchmark-static}
}
\end{floatrow}
\CenterFloatBoxes
\begin{floatrow}
\ttabbox[1.2\linewidth]{
    \centering
    \adjustbox{width=\linewidth}{
    \begin{tabular}{lllllllll}
    \toprule
    & Method & Mean & BiC-Catch & C-Swingup & C-Run & F-Spin & R-Easy & W-Walk\\
    \midrule
    \parbox[t]{1mm}{\multirow{6}{*}{\rotatebox[origin=c]{90}{Static setting}}} 
    & SAC & 76$\pm$6 & 109$\pm$9 & 167$\pm$19 & 77$\pm$15 & 4$\pm$3 & 75$\pm$9 & 24$\pm$2 \\
    & SAC+RAD & 113$\pm$12 & 96$\pm$14 & 272$\pm$11 & 21$\pm$15 & \bf 169$\pm$92 & \bf 93$\pm$6 & 25$\pm$1 \\
    & SAC+DrQ & 126$\pm$19 & 129$\pm$21 & 255$\pm$18 & 5$\pm$3 & \bf 259$\pm$107 & 82$\pm$11 & 26$\pm$1 \\
    & QT-Opt & 109$\pm$4 & 62$\pm$20 & 212$\pm$11 & 74$\pm$3 & 90$\pm$6 & \bf 109$\pm$7 & \bf 111$\pm$5 \\
    & QT-Opt+RAD & \bf 165$\pm$15 & \bf 172$\pm$12 & \bf 297$\pm$7 & \bf 130$\pm$7 & \bf 234$\pm$67 & \bf 94$\pm$16 & 63$\pm$3 \\
    & QT-Opt+DrQ & \bf 170$\pm$11 & \bf 169$\pm$25 & 283$\pm$5 & \bf 124$\pm$9 & \bf 266$\pm$51 & \bf 112$\pm$16 & 64$\pm$4 \\
    \midrule
    \parbox[t]{1mm}{\multirow{6}{*}{\rotatebox[origin=c]{90}{Dynamic setting}}} 
    & SAC & 86$\pm$3 & 102$\pm$24 & 175$\pm$6 & 57$\pm$4 & 1$\pm$0 & \bf 103$\pm$10 & 78$\pm$15 \\
    & SAC+RAD & 89$\pm$5 & 139$\pm$7 & 192$\pm$6 & 14$\pm$2 & \bf 63$\pm$24 & 93$\pm$6 & 31$\pm$2 \\
    & SAC+DrQ & 89$\pm$2 & \bf 185$\pm$20 & 185$\pm$2 & 16$\pm$1 & 15$\pm$7 & 101$\pm$6 & 31$\pm$1 \\
    & QT-Opt & 87$\pm$3 & 32$\pm$4 & 165$\pm$2 & \bf 71$\pm$4 & 28$\pm$11 & \bf 117$\pm$7 & \bf 112$\pm$4 \\
    & QT-Opt+RAD & \bf 103$\pm$3 & 132$\pm$20 & \bf 241$\pm$7 & 52$\pm$3 & 25$\pm$6 & \bf 105$\pm$10 & 64$\pm$2 \\
    & QT-Opt+DrQ & \bf 102$\pm$5 & 114$\pm$22 & \bf 243$\pm$5 & 54$\pm$2 & 26$\pm$5 & \bf 108$\pm$5 & 65$\pm$1 \\
    \bottomrule
    \end{tabular}}
}{
    \caption{Benchmark medium, $\beta_{\text{cam}}=\beta_{\text{rgb}}=0.2$, $b=8$}
    \label{table:medium}
}
\killfloatstyle\ffigbox[0.78\linewidth]{
    \centering
    \includegraphics[width=\linewidth]{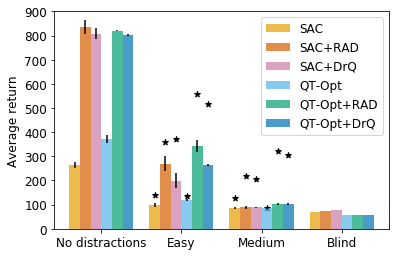}
}{%
    \caption{Benchmarks in the dynamic setting averaged over all tasks.}
    \label{fig:benchmark-dynamic}
}
\end{floatrow}
\end{table*}

In this section, we apply distractions during both training and evaluation. For the background, we vary the number of background videos during training, using the fully opaque distracting background ($\beta_{\text{bg}}=1$). Here we also look at generalization to unseen backgrounds during evaluation using the 30 videos from the test split of DAVIS 2017.

The results are shown in Figure~\ref{fig:train_with_distractions}. As before, performance drops with increasing distraction scale, which indicates how challenging it is for the agent to learn effectively in the presence of distractions. Training with distractions improves performance compared to the previous experiments for camera distractions and especially for background distractions, but not for color distractions (compare Fig.~\ref{fig:eval_with_distractions}a,b,c and Fig.~\ref{fig:train_with_distractions}a,b,c and note that for backgrounds the mixture weight is 1 in Fig.~\ref{fig:train_with_distractions}c,d). For background distractions, we can see that with more different training videos, the performance with these same videos decreases (Fig.~\ref{fig:train_with_distractions}c), while the performance on unseen videos increases and then levels off (see Fig.~\ref{fig:train_with_distractions}d).

Compared to the previous experiment, the static / dynamic setting appears to make a difference when training with distractions to camera pose and background. The dynamic setting (i.e. with moving cameras and video backgrounds, dashed lines) produces higher scores than the static setting (solid lines). This might result from allowing the agent to see a larger variety of distractions during training, i.e. a different distraction instance per frame instead of per episode. 

And contrary to the previous experiment, DrQ-based approaches are consistently outperforming SAC-based ones in all settings when training with distractions (compare blue/green to orange/pink lines in Fig.~\ref{fig:train_with_distractions}).

\begin{table}[t]
\floatsetup{captionskip=0cm}
\begin{floatrow}
\ttabbox[2\linewidth]{\centering
    \adjustbox{width=\linewidth}{
    \begin{tabular}{lllcccccc}
    \toprule
    & Difficulty & Method & Camera & Color & Backgr. & Product & & Benchmark\\
    \midrule
    \parbox[t]{1mm}{\multirow{8}{*}{\rotatebox[origin=c]{90}{Static setting}}} 
    & Easy & SAC+RAD & 0.79 & 0.74 & 0.41 & 0.24 & $>$ & 0.22 \\
    & Easy & SAC+DrQ & 0.81 & 1.01 & 0.41 & 0.33 & $>$ & 0.21 \\
    & Easy & QT-Opt+RAD & 0.88 & 0.92 & 0.60 & 0.48 & $>$ & 0.39 \\
    & Easy & QT-Opt+DrQ & 0.91 & 0.95 & 0.59 & 0.51 & $>$ & 0.37 \\
    & Medium & SAC+RAD & 0.58 & 0.64 & 0.46 & 0.17 & $>$ & 0.14 \\
    & Medium & SAC+DrQ & 0.55 & 0.76 & 0.42 & 0.18 & $>$ & 0.16 \\
    & Medium & QT-Opt+RAD & 0.57 & 0.92 & 0.54 & 0.28 & $>$ & 0.2 \\
    & Medium & QT-Opt+DrQ & 0.61 & 0.84 & 0.54 & 0.28 & $>$ & 0.21 \\
    \midrule
    \parbox[t]{1mm}{\multirow{8}{*}{\rotatebox[origin=c]{90}{Dynamic setting}}} 
    & Easy & SAC+RAD & 0.72 & 0.80 & 0.75 & 0.43 & $>$ & 0.32 \\
    & Easy & SAC+DrQ & 0.72 & 1.02 & 0.63 & 0.46 & $>$ & 0.25 \\
    & Easy & QT-Opt+RAD & 0.80 & 0.97 & 0.87 & 0.68 & $>$ & 0.42 \\
    & Easy & QT-Opt+DrQ & 0.84 & 0.93 & 0.83 & 0.65 & $>$ & 0.33 \\
    & Medium & SAC+RAD & 0.45 & 0.71 & 0.81 & 0.26 & $>$ & 0.11 \\
    & Medium & SAC+DrQ & 0.49 & 0.69 & 0.76 & 0.26 & $>$ & 0.11 \\
    & Medium & QT-Opt+RAD & 0.52 & 0.91 & 0.83 & 0.39 & $>$ & 0.13 \\
    & Medium & QT-Opt+DrQ & 0.54 & 0.90 & 0.79 & 0.38 & $>$ & 0.13 \\
    \bottomrule
    \end{tabular}}}{\caption{Interactions of distracting effects. Average scores w/ distractions relative to w/o distractions.}
    \label{table:interaction_of_distractions}}
\end{floatrow}
\end{table}

\subsection{A New Benchmark for Control from Pixels}

Here we combine all three distraction types.  We envision this combined setting as a new \emph{benchmark} for pixel-based RL that measures the ability to extract task-relevant information from visual input in the presence of visual distractions. To provide a set of competitive baselines for this benchmark, we evaluate the different combinations of SAC and QT-Opt with RAD and DrQ on this benchmark. 

To decide on the right values for the severity of distractions in the benchmark, we conducted experiments to generate an ``easy" and ``medium" difficulty for the tested methods.  We also added a ``blind" baseline to estimate lower bound of the performance in these tasks without seeing the relevant objects. In the easy setting, we use $\beta_{\text{cam}}=\beta_{\text{rgb}}=0.1$, and $b=4$ background videos. In the medium setting we use $\beta_{\text{cam}}=\beta_{\text{rgb}}=0.2$ and $b=8$.  
In the blind setting, we use the same parameters as the medium setting, but turn the camera backwards so that it cannot see any task-relevant information. All experiments are run with static as well as with dynamic distractions.

Figures~\ref{fig:benchmark-static} \& \ref{fig:benchmark-dynamic} show the average results across all tasks with no, easy, and medium distractions and for the blind benchmark. Detailed benchmark results can be found in Tables~\ref{table:default}, \ref{table:easy}, and \ref{table:medium}. The observations from these results are: 1) Sensitivity to distractions is task-dependent. The cheetah and walker tasks receive lower scores than ball in cup, cartpole, finger spin or reacher tasks. In the easy benchmark, the finger spin task works better in the static than in the dynamic setting, but for the reacher task it is flipped. 2) The performance degradation in these benchmarks is larger than the product of the individual performance reductions with the same parameters shown in Figure~\ref{fig:train_with_distractions}. Table
\ref{table:interaction_of_distractions} shows relative performance per distraction and reveals that their product is generally above the actual benchmark performance, which is also visualized in Figures~\ref{fig:benchmark-static} \& \ref{fig:benchmark-dynamic}. We find that the distractors have a compounding effect: combined, the distractors degrade performance more than individually.  This outcome is stronger in the dynamic than in the static setting. 3) In the medium benchmark, the static setting appears to be easier than the dynamic setting, where current methods only barely outperform the blind baseline experiment. Combined the easy and medium benchmarks should be a good metric for future research as they provide a lot of room for improvement, but still allow current methods to learn some meaningful behaviors. 4) The ranking of methods changes in the easy and medium benchmarks vs.\ no distractions, as QT-Opt methods now significantly outperform SAC-based methods. Random cropping is still essential to improve performance but does not ``solve'' these settings.

\section{Conclusion}
\label{sec:conclusion}

We have presented the \emph{Distracting Control Suite}, a new benchmark for pixel-based control in the presence of different types of visual distractions. We found that these distractions are challenging for current methods, especially when multiple distractions are applied at the same time. Between the methods that we compared, we found that random cropping was essential for good performance but DrQ did not outperform the simpler RAD approach. We also found that while SAC-based and QT-Opt-based methods perform similarly on the original DM Control benchmark, QT-Opt-based methods perform better in the presence of distractions, indicating that prior work on simpler environments might not transfer to more realistic settings. We hope that our benchmark\footnote{Code is available at \url{https://github.com/google-research/google-research/tree/master/distracting_control}} and analysis will facilitate progress towards algorithms that can efficiently handle the visual complexities of the real world.

\makeatletter
\providecommand{\doi}[1]{%
  \begingroup
    \let\bibinfo\@secondoftwo
    \urlstyle{rm}%
    \href{http://dx.doi.org/#1}{%
      doi:\discretionary{}{}{}%
      \nolinkurl{#1}%
    }%
  \endgroup
}

\bibliography{paper}  %

\begin{thebibliography}{10}

\bibitem{deepmindcontrolsuite2018}
Y.~Tassa, Y.~Doron, A.~Muldal, T.~Erez, Y.~Li, D.~de~Las~Casas, D.~Budden,
  A.~Abdolmaleki, J.~Merel, A.~Lefrancq, T.~Lillicrap, and M.~Riedmiller,
  ``Deep{Mind} control suite,'' tech. rep., DeepMind, Jan. 2018.

\bibitem{CuRL}
A.~Srinivas, M.~Laskin, and P.~Abbeel, ``{CURL}: Contrastive unsupervised
  representations for reinforcement learning,'' {\em arXiv preprint
  arXiv:2004.04136}, 2020.

\bibitem{DRQ}
I.~Kostrikov, D.~Yarats, and R.~Fergus, ``Image augmentation is all you need:
  Regularizing deep reinforcement learning from pixels,'' {\em arXiv preprint
  arXiv:2004.13649}, 2020.

\bibitem{RAD}
M.~Laskin, K.~Lee, A.~Stooke, L.~Pinto, P.~Abbeel, and A.~Srinivas,
  ``Reinforcement learning with augmented data,'' {\em arXiv preprint
  arXiv:2004.14990}, 2020.

\bibitem{jonschkowski2015learning}
R.~Jonschkowski and O.~Brock, ``Learning state representations with robotic
  priors,'' {\em Autonomous Robots}, vol.~39, no.~3, pp.~407--428, 2015.

\bibitem{zhang2020learning}
A.~Zhang, R.~McAllister, R.~Calandra, Y.~Gal, and S.~Levine, ``Learning
  invariant representations for reinforcement learning without
  reconstruction,'' {\em arXiv preprint arXiv:2006.10742}, 2020.

\bibitem{SAC}
T.~Haarnoja, A.~Zhou, P.~Abbeel, and S.~Levine, ``Soft actor-critic: Off-policy
  maximum entropy deep reinforcement learning with a stochastic actor,'' in
  {\em Proceedings of the 35th International Conference on Machine Learning},
  vol.~80, pp.~1861--1870, 2018.

\bibitem{QT}
D.~Kalashnikov, A.~Irpan, P.~Pastor, J.~Ibarz, A.~Herzog, E.~Jang, D.~Quillen,
  E.~Holly, M.~Kalakrishnan, V.~Vanhoucke, and S.~Levine, ``Scalable deep
  reinforcement learning for vision-based robotic manipulation,'' in {\em
  Proceedings of The 2nd Conference on Robot Learning}, vol.~87, pp.~651--673,
  2018.

\bibitem{DQN}
V.~Mnih, K.~Kavukcuoglu, D.~Silver, A.~A. Rusu, J.~Veness, M.~G. Bellemare,
  A.~Graves, M.~Riedmiller, A.~K. Fidjeland, G.~Ostrovski, {\em et~al.},
  ``Human-level control through deep reinforcement learning,'' {\em Nature},
  vol.~518, no.~7540, pp.~529--533, 2015.

\bibitem{pmlr-v97-hafner19a}
D.~Hafner, T.~Lillicrap, I.~Fischer, R.~Villegas, D.~Ha, H.~Lee, and
  J.~Davidson, ``Learning latent dynamics for planning from pixels,'' in {\em
  Proceedings of the 36th International Conference on Machine Learning},
  vol.~97, pp.~2555--2565, 2019.

\bibitem{pmlr-v28-levine13}
S.~Levine and V.~Koltun, ``Guided policy search,'' in {\em Proceedings of the
  30th International Conference on Machine Learning}, vol.~28, pp.~1--9, 2013.

\bibitem{SACX}
M.~Riedmiller, R.~Hafner, T.~Lampe, M.~Neunert, J.~Degrave, T.~van~de Wiele,
  V.~Mnih, N.~Heess, and J.~T. Springenberg, ``Learning by playing solving
  sparse reward tasks from scratch,'' in {\em Proceedings of the 35th
  International Conference on Machine Learning}, vol.~80, pp.~4344--4353, 2018.

\bibitem{InfiniteReplay}
S.~Cabi, S.~Gómez~Colmenarejo, A.~Novikov, K.~Konyushkova, S.~Reed, R.~Jeong,
  K.~Zolna, Y.~Aytar, D.~Budden, M.~Vecerik, O.~Sushkov, D.~Barker, J.~Scholz,
  M.~Denil, N.~de~Freitas, and Z.~Wang, ``Scaling data-driven robotics with
  reward sketching and batch reinforcement learning,'' in {\em Proceedings of
  Robotics: Science and Systems}, 2019.

\bibitem{SadeghiL16}
F.~Sadeghi and S.~Levine, ``{CAD2RL}: Real single-image flight without a single
  real image,'' in {\em Proceedings of Robotics: Science and Systems}, 2017.

\bibitem{tobin2017domain}
J.~Tobin, R.~Fong, A.~Ray, J.~Schneider, W.~Zaremba, and P.~Abbeel, ``Domain
  randomization for transferring deep neural networks from simulation to the
  real world,'' in {\em 2017 IEEE/RSJ International Conference on Intelligent
  Robots and Systems (IROS)}, pp.~23--30, IEEE, 2017.

\bibitem{rubikscube}
I.~Akkaya, M.~Andrychowicz, M.~Chociej, M.~Litwin, B.~McGrew, A.~Petron,
  A.~Paino, M.~Plappert, G.~Powell, R.~Ribas, {\em et~al.}, ``Solving rubik's
  cube with a robot hand,'' {\em arXiv preprint arXiv:1910.07113}, 2019.

\bibitem{lecun1998gradient}
Y.~LeCun, L.~Bottou, Y.~Bengio, and P.~Haffner, ``Gradient-based learning
  applied to document recognition,'' {\em Proceedings of the IEEE}, vol.~86,
  no.~11, pp.~2278--2324, 1998.

\bibitem{zhang2018natural}
A.~Zhang, Y.~Wu, and J.~Pineau, ``Natural environment benchmarks for
  reinforcement learning,'' {\em arXiv preprint arXiv:1811.06032}, 2018.

\bibitem{antonova2020benchmarking}
R.~Antonova, S.~Devlin, K.~Hofmann, and D.~Kragic, ``Benchmarking unsupervised
  representation learning for continuous control,'' in {\em Robotics
  Retrospectives Workshop at RSS}, 2020.

\bibitem{Hansen2020}
N.~Hansen, Y.~Sun, P.~Abbeel, A.~A. Efros, L.~Pinto, and X.~Wang,
  ``Self-supervised policy adaptation during deployment,'' {\em arXiv preprint
  arXiv:2007.04309}, 2020.

\bibitem{DAVIS}
J.~Pont-Tuset, F.~Perazzi, S.~Caelles, P.~Arbel\'aez, A.~Sorkine-Hornung, and
  L.~{Van Gool}, ``The 2017 {DAVIS} challenge on video object segmentation,''
  {\em arXiv:1704.00675}, 2017.

\bibitem{rubinstein1999cross}
R.~Rubinstein, ``The cross-entropy method for combinatorial and continuous
  optimization,'' {\em Methodology and computing in applied probability},
  vol.~1, no.~2, pp.~127--190, 1999.

\bibitem{ba2016layer}
J.~L. Ba, J.~R. Kiros, and G.~E. Hinton, ``Layer normalization,'' {\em arXiv
  preprint arXiv:1607.06450}, 2016.

\end{thebibliography}
\bibliographystyle{ieeetr}

\end{document}